\documentclass{article}
\usepackage{amsmath}
\usepackage{amssymb}
\usepackage{cim2026}
\usepackage{times}
\usepackage{ifpdf}
\usepackage[english]{babel}
\usepackage{booktabs}
\usepackage{microtype}
\usepackage{url}
\usepackage{subcaption}
\usepackage{tikz}
\usepackage{float}
\usetikzlibrary{arrows.meta,positioning,fit,backgrounds}

\newenvironment{metaphorbox}{\footnotesize\par\noindent}{\par}
\newcommand{\poeticlabel}{\textup{\textsc{Metaphor.}}\ }

\def\papertitle{MEMORY AS TRANSFORMATION: LETHE, A SELF-REFERENTIAL GAN-INSPIRED ARCHITECTURE}
\def\firstauthor{Francesco Vitucci}
\def\secondauthor{Anthony Di Furia}
\def\thirdauthor{Francesco Scagliola}

\ifpdf
  \usepackage{graphicx}
  \graphicspath{{img/fixed/}{img/circular/}}
  \usepackage[pdftex,
    pdftitle={\papertitle},
    pdfauthor={\firstauthor},
    bookmarksnumbered,
    pdfstartview=XYZ
  ]{hyperref}
\else
  \usepackage[dvips]{epsfig,graphicx}
  \usepackage[dvips,bookmarksnumbered,pdfstartview=XYZ]{hyperref}
\fi

\hypersetup{colorlinks,citecolor=black,filecolor=black,linkcolor=black,urlcolor=black}

\title{\papertitle}

\threeauthors
  {\firstauthor}{\footnotesize Conservatorio di Musica ``N. Piccinni'' di Bari \\ {\tt\footnotesize \href{mailto:francescovitucci1@gmail.com}{francescovitucci1@gmail.com}}}
  {\secondauthor}{\footnotesize Conservatorio di Musica ``N. Piccinni'' di Bari \\ {\tt\footnotesize \href{mailto:anthonydifuria.sound@gmail.com}{anthonydifuria.sound@gmail.com}}}
  {\thirdauthor}{\footnotesize Conservatorio di Musica ``N. Piccinni'' di Bari \\ {\tt\footnotesize \href{mailto:francesco.scagliola@gmail.com}{francesco.scagliola@gmail.com}}}
\begin{document}
\maketitle

\abstract
LETHE (Latent-parameter Evolution with Temporal Hierarchical quasi-Equilibrium) is a self-referential sonic-oblivion system implemented in SuperCollider. It adopts the formal vocabulary of Generative Adversarial Networks in a closed configuration without external datasets or supervision after initialization. Audio is processed by a 3$\times$3 mixing matrix built around two delay lines; its nine coefficients and two delay times evolve through the interaction of a five-feature linear discriminator and a random-perturbation optimizer analogous to single-sample REINFORCE. The discriminator compares current energy behavior with an archive of the initial state and guides parameter updates. Circular, fixed, and live sources can be mixed independently. Across fixed and circular sessions with an ablation control, the active generator is necessary for parametric evolution ($\Delta c_{22}=0.000$ in all 15 ablation sessions). Situated in the tradition of self-referential electroacoustic music, LETHE delegates the sonic outcome to an adaptive closed loop whose parametric space is defined by the composer.
\endabstract

\section{Introduction}\label{sec:introduzione}

In 1969 Alvin Lucier enters a room, records his own voice, plays the recording back in the same space, and re-records the result \cite{lucier:1969}. At each iteration the resonant frequencies of the room amplify some components and attenuate others, until every trace of the original linguistic content is erased. \textit{I Am Sitting in a Room} is a system that iteratively transforms its own input through the physical properties of the space: the transformation is deterministic, passive, invariant. The space does not respond to the sound; the sound responds to the space.

LETHE extends this logic by introducing an explicit and formally describable mechanism of agency. The Feedback Delay Network (FDN) replaces the physical space, while an adaptive control system actively modifies the transformation parameters at each cycle. Adopting the vocabulary of Generative Adversarial Networks (GANs) \cite{goodfellow:2014} as a descriptive tool for a non-differentiable system, LETHE realizes a simplified configuration: an energetic discriminator evaluates the coherence of the current state with respect to the system's initial behavior (its ``sonic DNA''), guiding a parameter-update mechanism (generator) that responds to this evaluation.

This self-referential architecture is situated in the tradition of ecosystemic signal processing \cite{discipio:2003} and of distributed agency \cite{hayles:1999, lewis:2000}: the composer defines the space of parametric possibilities, delegating the sonic outcome to the emergent interaction among the components. The modularity of the system furthermore allows the energetic evaluation function to be replaced with other metrics without altering the feedback logic of the generator.

This contribution documents the architecture of the system and the numerical validation of the adversarial mechanism through controlled experimental sessions, referring to the project repository for the full details of the SuperCollider implementation.

\section{Context: Classical GAN and Electroacoustic Tradition}\label{sec:contesto}

The framework of Generative Adversarial Networks (GANs), introduced by Goodfellow et al.\ \cite{goodfellow:2014}, describes a system composed of two competing neural networks: a generator $G$ that produces synthetic samples and a discriminator $D$ that compares them with samples drawn from a real distribution $p_{data}$. The objective function is formulated as a minimax game:

{\small\begin{equation}
\min_G \max_D \; \mathbb{E}_{x \sim p_{data}}[\log D(x)] + \mathbb{E}_{z \sim p_z}[\log(1 - D(G(z)))]
\end{equation}}

The theoretical equilibrium is reached when $D(x) = 0.5$ for every $x$, a condition known as the Nash equilibrium \cite{goodfellow:2014}{\cite{arjovsky:2017}}. The standard framework presupposes an external dataset as the real reference distribution, a latent space $z$ from which the generator samples, and a sharp separation between the real and the generated distributions.

Equation~(1) is a reference against which LETHE's simplification is defined, not an objective that it implements. The external dataset is replaced by the system's initial energetic behavior; the generator becomes a parameter-update mechanism; and the FDN parameters themselves, rather than samples drawn from a latent space, are optimized. The discriminator evaluates energetic coherence with the initial reference instead of classifying real and synthetic samples. LETHE therefore preserves the dynamic relation between evaluation and update, mediated by a scalar reward, in a non-differentiable real-time audio context \cite{goodfellow:2014}. Because the reference ``DNA'' is produced by the system during warm-up, the configuration also recalls autopoietic self-production \cite{maturana:1980}.

The tradition of self-referential electroacoustic music offers an equally pertinent artistic reference. In \textit{I Am Sitting in a Room} (1969), Alvin Lucier \cite{lucier:1969} records his own voice in a physical space, plays the recording back in the same space, and re-records the result iteratively, until the resonant frequencies of the room progressively amplify certain spectral components, erasing every trace of the original linguistic content. The transformation is deterministic and passive; the physical space acts as an invariant filter without responding to the sound that passes through it.

In LETHE the logic is analogous, but the process acquires agency: the Feedback Delay Network replaces the physical space, and the adaptive control system actively decides how to modify the transformation parameters at each cycle. David Tudor \cite{tudor:1968} and Gordon Mumma {\cite{mumma:1967}}, with their electronic feedback systems in the 1960s, and Agostino Di Scipio with the concept of ecosystemic signal processing \cite{discipio:2003} explored adjacent territories, while Nicolas Collins \cite{collins:2006}{\cite{collins:2021}} extensively documented this tradition in the practice of hardware hacking and feedback circuits. {Sanfilippo and} {Valle's analytical framework} {for feedback systems} {\cite{sanfilippo_valle:2013} offers a} {formal complement to} {this practice-based lineage.} LETHE positions itself at the intersection of these traditions, using the formal vocabulary of GANs to describe a process that experimental music has known empirically for decades.

\section{System Architecture}\label{sec:architettura}

\subsection{General Overview}\label{subsec:panoramica}

LETHE is organized as a closed loop with six main blocks. The sound material is recorded into a circular buffer {(by default 10~s} {long; see} {Section~\ref{subsec:buffer})}, processed by a 3$\times$3 Feedback Delay Network (FDN) whose parameters evolve autonomously, measured by three envelope followers, evaluated by a linear discriminator, modified by a random-perturbation parameter optimizer inspired by the single-sample REINFORCE \cite{williams:1992} scheme, and rewritten into the buffer for the next cycle.

The passages labeled ``Metaphor'' state the conceptual origin of design choices formalized in the technical sections.

\begin{metaphorbox}
{\poeticlabel}Just as in the waters of Lethe the soul would lose the memory of its previous lives before reincarnating, the buffer of LETHE continuously overwrites its own sonic history, while the system struggles to maintain a recognizable identity in the flow of transformation.
\end{metaphorbox}

\begin{figure}[htb]
 \centerline{%
 \scalebox{0.9}{%
 \begin{tikzpicture}[
  node distance=6pt,
  block/.style={rectangle, rounded corners=4pt, draw, fill=gray!10,
                text width=4.8cm, align=center, minimum height=28pt,
                font=\footnotesize},
  arrow/.style={-{Stealth[length=5pt]}, thick}
]
\node[block] (buf)  {Circular buffer\\{\scriptsize\texttt{genBuf} / \texttt{frozenBuf}}};
\node[block, below=of buf] (fdn)  {FDN 3$\times$3\\{\scriptsize matrix $c_{ij}$, delay lines}};
\node[block, below=of fdn] (ef)   {Envelope followers\\{\scriptsize $e_1$, $e_2$, $e_3$ @ 10 Hz via OSC}};
\node[block, below=of ef]  (disc) {Discriminator\\{\scriptsize perceptron, 5 features, sigmoid}};
\node[block, below=of disc](gen)  {Perturbation optimizer\\{\scriptsize reward, baseParams, perturbation}};
\node[block, below=of gen] (coup) {Coupling + $\rho$\\{\scriptsize hierarchy, spectral radius}};
\draw[arrow] (buf)  -- (fdn);
\draw[arrow] (fdn)  -- (ef);
\draw[arrow] (ef)   -- (disc);
\draw[arrow] (disc) -- (gen);
\draw[arrow] (gen)  -- (coup);
\draw[arrow] (coup.south) -- ++(0,-10pt) -| ([xshift=18pt]buf.east) |- (buf.east);
\end{tikzpicture}}}
 \caption{General scheme of LETHE. The closed loop receives no external signals after the initialization phase.}
 \label{fig:diagramma1}
\end{figure}
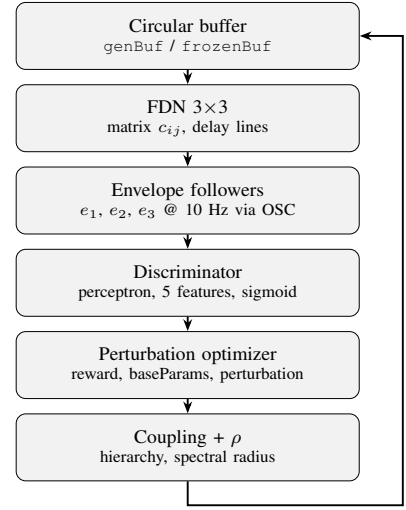

The general block diagram is shown in Figure~\ref{fig:diagramma1}.

\begin{metaphorbox}
{\poeticlabel}The circular buffer is the river carrying and overwriting its history; \texttt{frozenBuf} is its initial photograph. The FDN is the riverbed, the envelope followers measure its level, and the discriminator retains a partial memory of its initial identity. The optimizer shifts the bed, while coupling and spectral-radius control act as the laws and embankments that constrain the flow. Circular, fixed, and live sources correspond respectively to transformed memory, preserved memory, and new material entering the system.
\end{metaphorbox}

\subsection{The Detailed Processing Chain}\label{subsec:catena}

The processing chain comprises three closed-loop blocks. The audio server records or reads the source, applies the FDN and delay lines, and reinjects the output into the circular buffer. At 10~Hz, the measurement bridge sends three envelope values via OSC and derives five normalized features. The client populates the DNA archive, evaluates and trains the discriminator, computes the reward and perturbation update, applies coupling and the condition $\rho(\mathbf{F})<0.99$, and returns the parameters to the synth. The complete SuperCollider implementation \cite{mccartney:2002} is available in the project repository.

\section{Implementation}\label{sec:implementazione}

\subsection{Buffer Allocation and Input Recording}\label{subsec:buffer}

The system allocates in memory a mono buffer of length $N = \lfloor f_s \cdot T_{buf} \rfloor$ samples. A second identical buffer, named \texttt{frozenBuf}, receives a copy of the contents of the main buffer immediately after the initial recording and is never overwritten. The complete implementation of buffer allocation and management is available in the file \texttt{02\_synth.scd} of the repository.

In \texttt{\textbackslash mic} {acquisition} mode the system records the incoming audio signal via \texttt{RecordBuf.ar}. This is the only moment in which LETHE receives information from outside.

\subsection{The 3$\times$3 Feedback Delay Network}\label{subsec:fdn}

The FDN constitutes the core of the synthesis. The input signal $g(t)$ and the delayed outputs $d_1(t)$, $d_2(t)$ coming from the two delay lines are combined according to the linear transformation of the $3\times3$ FDN matrix, generating the three outputs $y_1$, $y_2$, $y_3$. Adopting the standard matrix convention in which $c_{ij}$ denotes the contribution of input $j$ to output $i$ --- that is, row $i$ corresponds to output $i$ and column $j$ to input $j$ ---:

{\small\begin{equation}
\begin{bmatrix} y_1(t) \\ y_2(t) \\ y_3(t) \end{bmatrix}
=
\begin{bmatrix} c_{11} & c_{12} & c_{13} \\ c_{21} & c_{22} & c_{23} \\ c_{31} & c_{32} & c_{33} \end{bmatrix}
\begin{bmatrix} g(t) \\ d_1(t) \\ d_2(t) \end{bmatrix}
\label{eq:fdn}
\end{equation}}

The nine coefficients $c_{ij}$ and the two delay times $\tau_1, \tau_2$ constitute the eleven parameters controlled by the adaptive system. In the standard matrix convention the three rows correspond to the three outputs:
\begin{itemize}
  \setlength{\itemsep}{0pt}
  \setlength{\parskip}{0pt}
  \setlength{\parsep}{0pt}
  \setlength{\topsep}{0pt}
  \setlength{\partopsep}{0pt}
  \item \textbf{Row 1} ($c_{11}, c_{12}, c_{13}$): governs $y_1$, the output to the loudspeakers. $c_{11}$ quantifies the contribution of the direct signal; $c_{12}$ and $c_{13}$ the feedback contributions of the two delays to the direct output.
  \item \textbf{Row 2} ($c_{21}, c_{22}, c_{23}$): governs $y_2$, the input of the first delay. $c_{21}$ = routing of the fresh signal toward $d_1$; $c_{22}$ = self-feedback of $d_1$ (the most important for the vitality of the delay); $c_{23}$ = cross-feedback from $d_2$ toward $d_1$.
  \item \textbf{Row 3} ($c_{31}, c_{32}, c_{33}$): governs $y_3$, the input of the second delay. $c_{31}$ = routing of the fresh signal toward $d_2$; $c_{32}$ = cross-feedback from $d_1$ toward $d_2$; $c_{33}$ = self-feedback of $d_2$.
\end{itemize}

The outputs of the two delay lines are determined by the times $\tau_1, \tau_2 \in [\tau_{min}, \tau_{max}]$, parameters optimized in real time by the generator. {In the present} {implementation $\tau_{min}=1$~ms} {and $\tau_{max}=100$~ms,} {corresponding to comb-filter} {first-resonance spacings ranging} {from about 1~kHz} {down to 10~Hz.} The FDN structure for artificial reverberation synthesis is well established in the literature \cite{schroeder:1962, jot:1991, rocchesso:1997, valimaki:2012, moorer:1979, smith:2010}. Time-varying-matrix FDNs have been proposed for adaptive artificial reverberation synthesis \cite{schlecht:2015}, with deterministic modulation oriented toward acoustic objectives. LETHE adopts an analogous FDN structure but replaces the deterministic modulation with a stochastic optimization process guided by the discriminator.

The internal feedback loop is implemented via \texttt{LocalIn} \texttt{LocalOut}, a standard SuperCollider mechanism that introduces a one-processing-block delay to prevent infinite loops in the signal graph \cite{mccartney:2002}. The complete implementation of the \texttt{\textbackslash lethe\_fdn} SynthDef is available in the repository.

Each parameter update is linearly interpolated to prevent audible discontinuities in the audio signal. The delay times are kept internally in the normalized space $[0,1]$ and remapped to the real range $[\tau_{min}, \tau_{max}]$ via \texttt{linlin} in the \texttt{applyPerturb} function.

The rewriting of the buffer occurs by summing the outputs of the network and scaling the result by the parameter $\alpha_{buf}$ (\texttt{feedbackGBuffer}). The signal is also filtered by a first-order high-pass filter to remove the direct-current (DC) components which, accumulating at each cycle, would compromise the stability of the system \cite{zolzer:2008}.

\subsection{Measurement and OSC Communication}\label{subsec:osc}

Three envelope followers measure the amplitude of the signal at three points of the FDN: the direct output $y_1$, the first delay $d_1$, the second delay $d_2$. The three values $[e_1, e_2, e_3]$ are sent to the SuperCollider client process via the OSC protocol \cite{wright:1997} at frequency $f_{ctrl} = 10$ Hz. This cadence defines the temporal resolution of the entire control system: every discriminator decision and every generator update occurs on this 100 ms grid.

\subsection{The Discriminator}\label{subsec:discriminatore}

The discriminator is a single perceptron \cite{rosenblatt:1958} with five input features and a sigmoid activation function. This choice is deliberate: a minimally parameterized discriminator, whose behavior is interpretable through its own weights, privileges the transparency of the mechanism over discriminative capacity (cf.\ Section~\ref{sec:discussione}).

The five features $\hat{f}_i$ are computed from the raw values of the envelope followers:

{\small\begin{align}
\hat{f}_1 &= \frac{e_1}{\max(e_1 + e_2 + e_3,\; \varepsilon)} \\
\hat{f}_2 &= \frac{e_2}{\max(e_1 + e_2 + e_3,\; \varepsilon)} \\
\hat{f}_3 &= \frac{e_3}{\max(e_1 + e_2 + e_3,\; \varepsilon)} \\
\hat{f}_4 &= e_1 + e_2 + e_3 \\
\hat{f}_5 &= \frac{\hat{f}_4 - \bar{f}_4^{\text{rec}}}{\max(\bar{f}_4^{\text{rec}},\; \varepsilon)}
\end{align}}

where $\varepsilon = 10^{-3}$ avoids divisions by zero and $\bar{f}_4^{\text{rec}}$ is the moving average of $\hat{f}_4$ computed over a circular window of 128 samples. The features $\hat{f}_1, \hat{f}_2, \hat{f}_3$ capture the relative distribution of energy among the three outputs, invariant with respect to the absolute level. The feature $\hat{f}_4$ captures the absolute level. The feature $\hat{f}_5$ captures the energetic trend with respect to the recent history.

Each raw feature $\hat{f}_i$ is adaptively normalized, yielding the normalized feature $\tilde{f}_i$, with an exponential moving-average update \cite{ioffe:2015}{\cite{welford:1962}} {Let $\mu_i$ and} {$\sigma_i^2$ denote the} {running mean and variance} {of feature $i$,} {updated online via} {exponential moving average} {with rate $\alpha_f$}. Setting $\delta_i = \hat{f}_i - \mu_i$ (computed before the update of $\mu_i$):

{\small\begin{align}
\mu_i &\leftarrow \mu_i + \alpha_f \delta_i \\
\sigma_i^2 &\leftarrow \sigma_i^2(1-\alpha_f) + \alpha_f \delta_i^2 \\
\tilde{f}_i &= \text{clip}\!\left(\frac{\hat{f}_i-\mu_i}{\sqrt{\sigma_i^2+\varepsilon}},\,-3,\,3\right)
\end{align}}
where $\mu_i$ and $\sigma_i^2$ in the last line are the already-updated values.

The forward pass of the discriminator computes the score $s \in (0,1)$ from the normalized features $\tilde{f}_i$:

{\small\begin{equation}
s = \sigma\!\left(\sum_{i=1}^{5} w_i \tilde{f}_i + b\right) = \frac{1}{1+e^{-(\mathbf{w}^T\tilde{\mathbf{f}}+b)}}
\label{eq:discriminatore}
\end{equation}}

Values of $s$ close to 1 indicate that the current sound is coherent with the DNA of the archive; values close to 0 indicate divergence.

The training of the discriminator minimizes the Binary Cross-Entropy (BCE) over two samples per cycle: the current state of the system (treated as a \textit{fake} sample) and a random sample from the DNA archive (treated as a \textit{real} sample). The loss is:

{\small\begin{equation}
\mathcal{L}_D = -\left[\log D(\tilde{\mathbf{f}}^{real}) + \log(1 - D(\tilde{\mathbf{f}}^{fake}))\right]
\end{equation}}

The gradient with respect to the weights $w_i$ and the bias $b$ is:

{\small\begin{align}
\frac{\partial \mathcal{L}_D}{\partial w_i} &= \tilde{f}_i^{fake} \cdot s_{fake} - \tilde{f}_i^{real} \cdot (1-s_{real}) \\
w_i &\leftarrow w_i - \eta_D \cdot \frac{\partial \mathcal{L}_D}{\partial w_i} \\
\frac{\partial \mathcal{L}_D}{\partial b} &= s_{fake} - (1 - s_{real}) \\
b &\leftarrow b - \eta_D \cdot \frac{\partial \mathcal{L}_D}{\partial b}
\end{align}}

where $\eta_D$ is the learning rate of the discriminator. Weights and bias are confined to the interval $[-8, 8]$ to prevent gradient explosion. The complete implementation of the training function \texttt{\textasciitilde dTrain} is available in the file \texttt{03\_gan.scd} of the repository.

\subsection{The DNA Archive}\label{subsec:archivio}

The archive is a circular buffer of $A$ slots (default $A=100$), each containing a vector of five normalized features $\tilde{f}_i$. During the warm-up phase, the archive is populated with the observed vectors (for a duration equal to one complete buffer cycle, or manually configurable in \texttt{\textbackslash live} {acquisition} mode). At the end of this phase, the archive constitutes the system's ``sonic DNA'', that is, the operational representation of the distribution $p_{data}$ in the GAN analogy.

During training, the archive is updated conditionally: a new sample is added only if its score $s > \theta_{arch}$. This policy keeps the DNA as a collection of states coherent with the initial behavior, avoiding contamination with distant states. Since in practice the mean score turns out to be lower than $\theta_{arch}$ (cf.\ Section~\ref{sec:validazione}), the archive updates rarely, in effect preserving the reference distribution captured during warm-up.

{A methodological note} {is due here: the} {running mean $\mu_i$} {and variance $\sigma_i^2$} {used for feature} {normalization (Section~\ref{subsec:discriminatore}) are} {never frozen after} {warm-up and keep} {updating on every} {cycle. Each archive} {vector is stored} {already normalized under} {the statistics in} {effect at the} {moment of insertion;} {as $\mu_i,\sigma_i^2$ subsequently} {drift, the archive} {and the live} {discriminator input may} {end up expressed} {in slightly different} {normalization frames. This} {is a limitation} {of the current} {implementation that we} {have not verified} {to be negligible,} {and is left} {as a direction} {for future refinement} {(e.g.\ freezing normalization} {statistics after warm-up,} {or re-normalizing the} {archive on read).}

\subsection{The Generator: Parameter Update}\label{subsec:generatore}

The update of the FDN parameters plays the role of the generator in the GAN analogy: it responds to the discriminator's signal and pushes the parameters toward configurations that the discriminator judges coherent with the DNA. The mechanism uses random perturbations as estimators of the gradient of the reward, structurally analogous to the single-sample REINFORCE scheme \cite{williams:1992}, with which it shares the update structure (Eq.~\ref{eq:nes}): a randomly sampled perturbation, weighted by the reward signal centered on the baseline. The reference to Evolution Strategies \cite{wierstra:2008, salimans:2017} is conceptual in nature: NES uses the natural gradient over a population of candidate solutions, whereas LETHE operates on a single parametric configuration without requiring the differentiability of the system.

The single-sample design is not motivated by computational lightness but by architectural coherence: introducing a population of $K$ parallel configurations would require redefining which state is heard at the output and how to aggregate the $K$ scores --- choices incompatible with the unity of the observing subject. The system that evaluates is the same system that is modified.

Every $N_{upd}$ OSC cycles (default $N_{upd}=10$), the system computes the reward:

{\small\begin{equation}
r = \bar{s}_D - b_{base}
\end{equation}}

where $\bar{s}_D = \frac{1}{N_{upd}}\sum_k s_k$ is the mean of the $D$ scores over the last $N_{upd}$ cycles and $b_{base}$ is a baseline updated with an exponential moving average:

{\small\begin{equation}
b_{base} \leftarrow (1-\alpha_b) \cdot b_{base} + \alpha_b \cdot \bar{s}_D
\end{equation}}

The perturbation~$\varepsilon_j$ is held fixed throughout the $N_{upd}$-cycle window, so the reward is evaluated in the presence of the same perturbation used in the update---a necessary condition for the validity of the \textsc{spsa}{~\cite{spall:1992}} gradient estimate.

Subtracting the baseline reduces the variance of the estimated gradient. The update for each base parameter $\theta_j$ is:

{\small\begin{equation}
\theta_j \leftarrow \theta_j + \eta_G \cdot \varepsilon_j \cdot r
\label{eq:nes}
\end{equation}}

where $\varepsilon_j$ is the perturbation sampled for the current cycle and $\eta_G$ is the learning rate.

The perturbation is sampled with a two-level hierarchical mechanism that reflects the physical importance of each parameter:

{\small\begin{equation}
\varepsilon_j \sim \begin{cases} \mathcal{U}(-0.3,\,1.0) \cdot \sigma_j & j \in \{c_{22}, c_{33}\} \\ \mathcal{U}(-1.0,\,1.0) \cdot \sigma_j & \text{otherwise} \end{cases}
\end{equation}}

The positive bias on $c_{22}$ and $c_{33}$ --- the self-feedback coefficients of the two delay lines, the most important for the vitality of the system --- keeps the generator in a regime of predominantly positive feedback without forcing it: it is a weak prior, not a hard constraint. Since $b_{base}$ tracks $\bar{s}_D$ via \textsc{ema}, $\mathrm{E}[r]\approx 0$ (score~$D\approx 0.44$ across all sessions), so $\mathrm{E}[\Delta\theta_j]\approx 0$ regardless of $\mathrm{E}[\varepsilon_j]$. The complete implementation of the \texttt{\textasciitilde sampleNewPerturb} function is available in the repository.

\subsection{The Hierarchical Coupling}\label{subsec:accoppiamento}

After each update, the system propagates the changes among the parameters of the matrix according to three design-motivated relations, chosen to favor timbral diversification between the two delays and to reduce the risk of degenerate behaviors, not derived from a conservation principle intrinsic to the FDN structure. Let $\Delta c_{22}$ and $\Delta c_{33}$ be the variations of the self-feedback coefficients \textit{proposed by the generator} at the current cycle:

{\small\begin{align}
\text{Anti-correlation:} \quad & c_{33} \leftarrow c_{33} - \alpha_{AC}\,\Delta c_{22} \nonumber \\
                                & c_{22} \leftarrow c_{22} - \alpha_{AC}\,\Delta c_{33} \\
\text{Co-correlation:} \quad   & c_{23} \leftarrow c_{23} + \alpha_{CC}\,\Delta c_{32} \nonumber \\
                                & c_{32} \leftarrow c_{32} + \alpha_{CC}\,\Delta c_{23} \\
\text{Weak follow:} \quad     & c_{21} \leftarrow c_{21} + \alpha_{FF}\,\Delta c_{22} \nonumber \\
                                & c_{31} \leftarrow c_{31} + \alpha_{FF}\,\Delta c_{33}
\end{align}}

The anti-correlation between $c_{22}$ and $c_{33}$ creates asymmetry between the two delays: when one increases its self-feedback, the other decreases it. The co-correlation keeps the cross loop balanced. The weak follow of the inputs adapts the routing of the fresh signal to the dominant delay: $c_{21}$ and $c_{31}$ --- the shares of fresh signal that enter the first and the second delay path respectively --- weakly follow the respective self-feedbacks.

\subsection{Stability Control: Spectral Radius}\label{subsec:stabilita}

{Here ``spectral radius''} {denotes the largest-magnitude} {eigenvalue of the} {feedback matrix $\mathbf{F}$} {defined below, a} {linear-algebra notion unrelated} {to the frequency} {spectrum of the audio signal.} The stability of the FDN is determined primarily by the 2$\times$2 feedback submatrix:

{\small\begin{equation}
\mathbf{F} = \begin{bmatrix} c_{22} & c_{23} \\ c_{32} & c_{33} \end{bmatrix}
\end{equation}}

For this specific topology --- in which $y_2$ and $y_3$ constitute exclusively the feedback paths toward the delay lines --- the condition $\rho(\mathbf{F}) < 1$ on the spectral radius of the feedback block represents the theoretical stability condition of the closed loop in the approximation of exact delays \cite{jot:1991, horn:1985}{\cite{schlecht_habets:2017}}. For time-varying~$F$, \cite{schlecht:2015} show that instantaneous $\rho$ does not guarantee stability; $\rho < 0.99$ is therefore a heuristic safeguard, confirmed empirically across all~45 sessions. The spectral radius is $\rho(\mathbf{F}) = \max(|\lambda_1|, |\lambda_2|)$ where the eigenvalues are computed from the formula:

{\small\begin{equation}
\lambda_{1,2} = \frac{(c_{22}+c_{33}) \pm \sqrt{(c_{22}+c_{33})^2 - 4(c_{22}c_{33}-c_{32}c_{23})}}{2}
\end{equation}}

If the discriminant is negative (complex conjugate eigenvalues), $\rho = \sqrt{c_{22}c_{33}-c_{32}c_{23}}$. The choice of the spectral radius instead of the Frobenius norm --- a more conservative criterion --- is deliberate: the Frobenius norm would suppress configurations that are in fact stable \cite{jot:1991}.

If $\rho > 0.99$, the system applies a two-level hierarchical enforcement: first it scales $c_{32}$ and $c_{23}$ by a factor of 0.9; if $\rho$ remains above the threshold, it scales the entire block by a factor of $0.99/\rho$. The threshold $0.99 < 1$ introduces a safety margin with respect to the theoretical limit, compensating for the numerical imprecisions of floating-point arithmetic and the one-sample delay introduced by digital processing, which relaxes the stability condition with respect to the continuous-domain analysis.

\section{Project Structure}\label{sec:struttura}

\subsection{File Organization}\label{subsec:file}

The system is organized into five separate SuperCollider files for modularity and readability: \texttt{00\_main.scd} (entry point), \texttt{01\_config.scd} (configurable parameters), \texttt{02\_synth.scd} (FDN SynthDef and buffer management), \texttt{03\_gan.scd} (discriminator, optimizer, coupling, OSCdef), and \texttt{04\_log.scd} (CSV logging). The files are loaded in sequence via \texttt{load()}: each module initializes its own variables before loading the next, ensuring the correct dependency among the components.

\subsection{Availability of the Materials}\label{subsec:materiali}

The complete, versioned implementation of LETHE is openly available through the project repository~\cite{Vitucci_LETHE_2026}. It includes the SuperCollider source files (\texttt{00\_main.scd}, \texttt{01\_config.scd}, \texttt{02\_synth.scd}, \texttt{03\_gan.scd}, and \texttt{04\_log.scd}) and the audio examples discussed in Section~\ref{sec:sonoro}.

To support the reproducibility of the validation reported in Section~\ref{sec:validazione}, the SuperCollider configurations for the three experimental conditions (\textbf{fixed GAN}, \textbf{fixed ablation}, and \textbf{circular GAN}), together with the corresponding CSV datasets, are documented and deposited in the accompanying Zenodo archive~\cite{lethe_validation}.

\subsection{Source Mixer}\label{subsec:modalita}

LETHE provides three independent sound sources{, distinct from} {the \texttt{\textbackslash mic}/\texttt{\textbackslash file}/\texttt{\textbackslash live}} {acquisition modes of} {Section~4.1,} individually controllable via three amplitude parameters:

\begin{itemize}
\setlength{\itemsep}{0pt}
\setlength{\parskip}{0pt}
\setlength{\parsep}{0pt}
\setlength{\topsep}{0pt}
\setlength{\partopsep}{0pt}
\item \textbf{circular} (\texttt{ampCircular}): the circular buffer, rewritten by the FDN output at each cycle. It produces a progressive transformation of the sound material over time.
\item \textbf{fixed} (\texttt{ampFixed}): the \texttt{frozenBuf}, a permanent copy of the recorded material that is never overwritten. It produces a stable transformation: the adversarial mechanism operates on invariant sound material.
\item \textbf{live} (\texttt{ampLive}): direct input from the microphone or audio interface (\texttt{SoundIn.ar}). No buffer: the system processes in real time.
\end{itemize}

The three sources are summed with a smooth crossfade (\texttt{lag(1.0)}). The implementation of the mixer is available in the file \texttt{02\_synth.scd} of the repository.

This architecture allows any combination: circular only, fixed only, live only, or any mix of the three. The transition from one {source configuration} to another typically occurs without interruption of the sound.

\subsection{Configurable Parameters}\label{subsec:parametri}

All system parameters are centralized in \texttt{01\_config.scd}. Table~\ref{tab:parametri} summarizes the main parameters with their default values and their effect on the behavior of the system. The parameters \texttt{del1\_time} and \texttt{del2\_time} (delay times normalized in $[0,1]$, remapped to the real range $[\tau_{min}, \tau_{max}]$) complete the eleven FDN parameters controlled by the adaptive system but are not listed separately, as they are initialized by the warm-up and managed internally by the \texttt{applyPerturb} function. The value \texttt{auto} of \texttt{\textasciitilde warmUpCycles} in Table~\ref{tab:parametri} indicates that in \texttt{\textbackslash mic} and \texttt{\textbackslash file} {acquisition} modes the parameter is computed automatically by \texttt{02\_synth.scd} as $\lfloor T_{buf} \cdot 10 \rfloor$ cycles, overriding the value possibly set in \texttt{01\_config.scd}; in \texttt{\textbackslash live} {acquisition} mode the value from \texttt{01\_config.scd} is used directly.

\begin{table}[htb]
\begin{center}
\footnotesize
\setlength{\tabcolsep}{4pt}
\begin{tabular}{lcc}
\toprule
\textbf{Parameter} & \textbf{Default} & \textbf{Effect} \\
\midrule
\texttt{bufDur} & 10.0 s & Buffer and warm-up duration \\
\texttt{feedbackGBuffer} & 0.75 & Life of the system \\
\texttt{warmUpCycles} & auto & Initial observation cycles \\
\texttt{archiveSize} & 100 & DNA capacity \\
\texttt{archiveThreshold} & 0.55 & DNA selectivity \\
\texttt{dLR} & 0.005 & D learning rate \\
\texttt{gLR} & 0.5 & G update rate \\
\texttt{gUpdateEvery} & 10 & G update frequency \\
\texttt{coupAC} & 0.4 & Anti-correlation strength \\
\texttt{coupCC} & 0.6 & Co-correlation strength \\
\texttt{coupFF} & 0.25 & Follow strength \\

\bottomrule
\end{tabular}
\end{center}
\caption{Main configurable parameters of LETHE with default values.}
\label{tab:parametri}
\end{table}

The module \texttt{04\_log.scd} generates two CSV files at 10 Hz: \texttt{lethe\_log.csv} (all system parameters per cycle) and \texttt{lethe\_archive.csv} (snapshot of the archive at each update), allowing the complete reconstruction of the evolution of the session.

\section{Numerical Validation}\label{sec:validazione}

\subsection{Methodology and Results}\label{subsec:metodologia}

Forty-five experimental sessions were conducted: five types of input signal (white noise {(WN)}, a 2~Hz impulse train {(IMP)}, pink noise {(PN)}, square wave {(SQR)}, and sawtooth at 110~Hz {(SAW)}) $\times$ three conditions (fixed GAN, fixed ablation, circular GAN) $\times$ three repetitions, each of 180~s. {The three conditions} {are defined as follows:} {in \textbf{fixed GAN}} {the adaptive mechanism} {(discriminator and generator)} {is fully active} {on the frozen-buffer} {source; in \textbf{fixed ablation}} {the generator learning} {rate is set} {to zero ($\eta_G = 0$),} {disabling parameter updates} {while the discriminator} {keeps running, on} {the same frozen-buffer} {source; in \textbf{circular GAN}} {the adaptive mechanism} {is fully active} {on the circular} {(self-overwriting) buffer source.} The complete validation is documented in the technical report~\cite{lethe_validation}.

Table~\ref{tab:confronto} summarizes the aggregate metrics. The key result is that the active generator is a \emph{necessary condition} for parametric evolution: in all~15 ablation sessions ($\eta_G = 0$) $c_{22}$ remains unchanged ($\Delta c_{22} = 0.000$), while in the 15~Fixed \textsc{gan} sessions $\Delta c_{22} = {+0.102 \pm 0.067}$ ($t(14) = 5.687$, $p < 0.0001$; Wilcoxon $p = 0.0002$). The direction of variation is positive in 14~out of~15 \textsc{gan} sessions (sign test: $p < 0.001$), ruling out numerical drift, buffer artifacts, and sampling bias as alternative explanations.

\begin{table}[H]
\begin{center}
\footnotesize
\setlength{\tabcolsep}{3pt}
\begin{tabular}{lccc}
\toprule
\textbf{Metric} & \textbf{Fixed GAN} & \textbf{Ablation} & \textbf{Circular GAN} \\
\midrule
$\mu$ score $D$                    & 0.441              & 0.446             & 0.433 \\
$\sigma$ reward (WN)               & 0.061              & 0.065             & \textbf{0.089} \\
Corr base $c_{22}/c_{33}$ (WN)     & $-$0.995           & n.a.              & $-$0.969 \\
$\Delta$ base $c_{22}$             & \textbf{$+$0.102 $\pm$ 0.067} & \textbf{0.000} & $+$0.101 $\pm$ 0.081 \\
\bottomrule
\end{tabular}
\end{center}
\caption{Comparison among the three conditions. $\mu$ score $D$ and $\Delta$ base $c_{22}$: aggregate mean over $n=15$ sessions (5 signals $\times$ 3 repetitions), common window of 180~s. $\sigma$ reward and correlation $c_{22}/c_{33}$: representative WN session.}
\label{tab:confronto}
\end{table}

\begin{figure}[H]
\centerline{\includegraphics[width=\columnwidth]{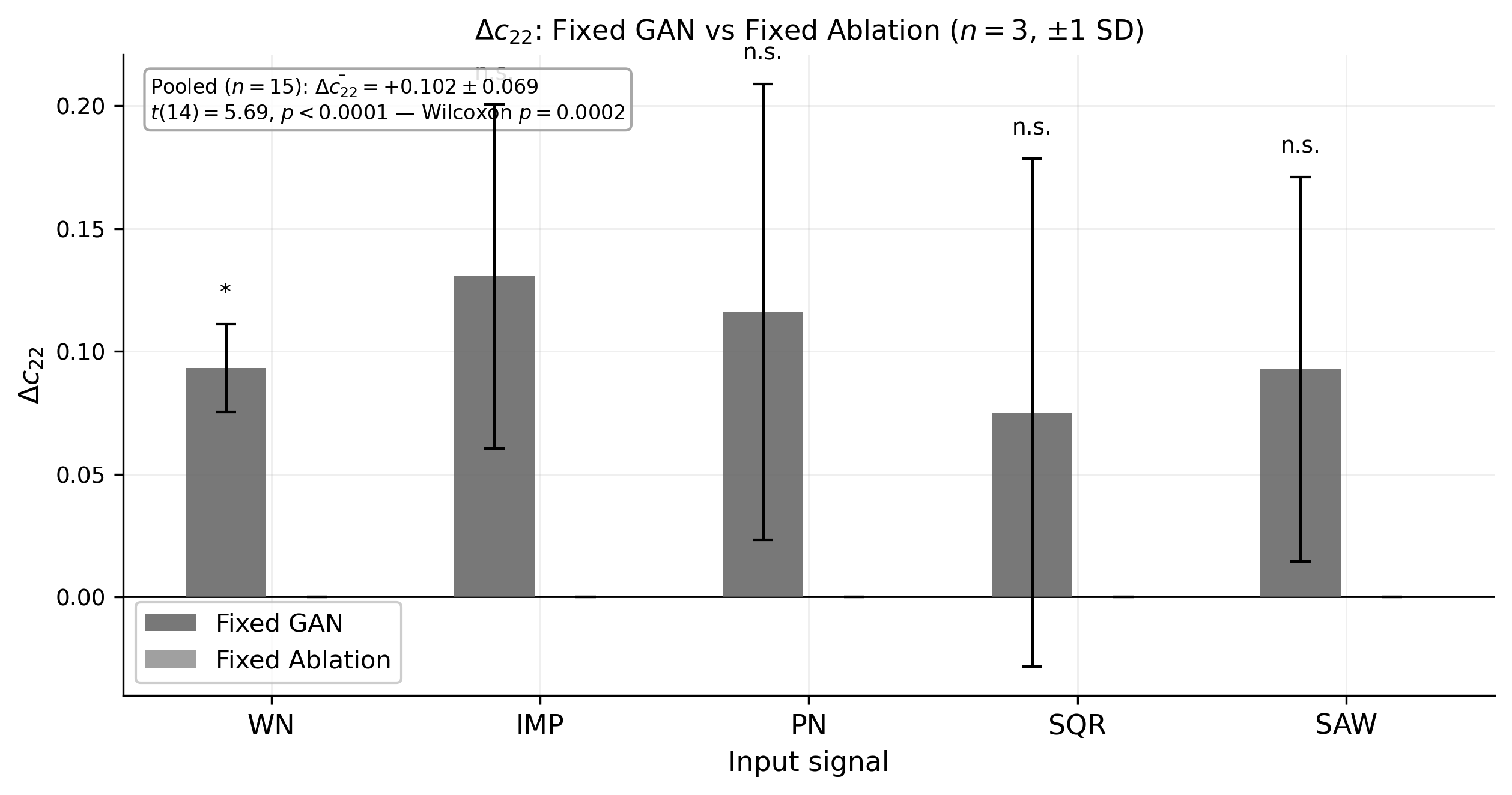}}
\caption{$\Delta c_{22}$: Fixed GAN vs.\ Fixed Ablation for all signals ($n=3$, error bars $=\pm 1$ std.\ dev.). The marker~* indicates $p<0.05$ in the one-sample $t$-test (df$=2$). Ablation at zero for all signals{: the Fixed} {Ablation bars are} {not visually distinguishable} {from the axis} {because their height} {is exactly zero} {by construction ($\eta_G=0$),} {not because data} {is missing}. The inset reports the aggregate statistics ($n=15$).}
\label{fig:delta_c22}
\end{figure}

\begin{figure}[H]
\centerline{\includegraphics[width=\columnwidth]{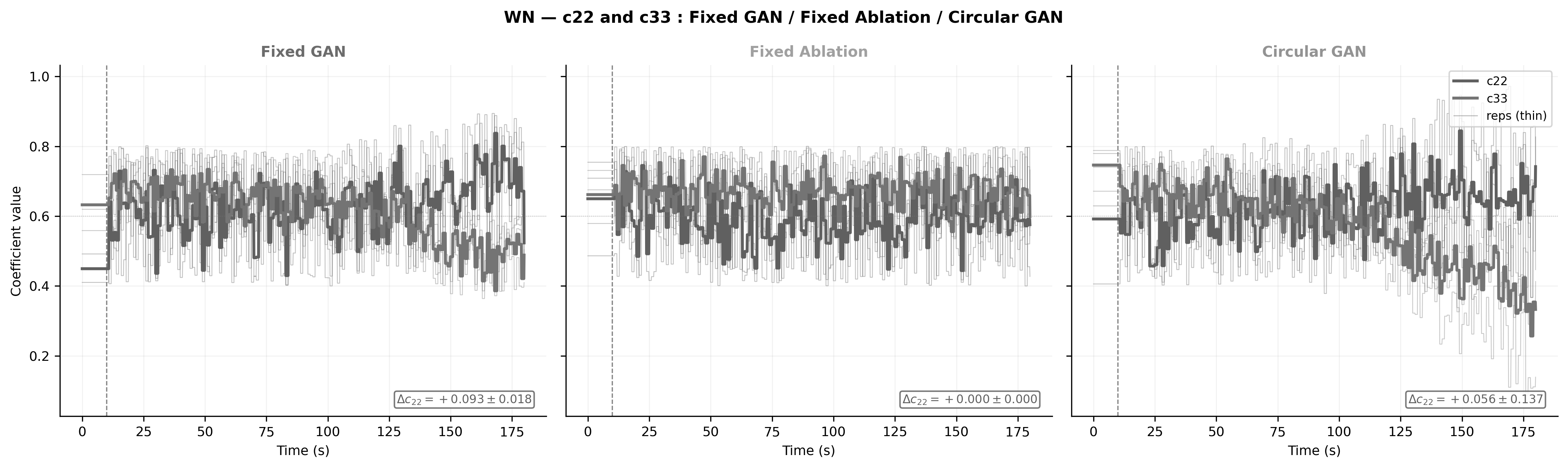}}
\caption{Evolution of  {$c_{22}$ and $c_{33}$} {(distinguished by line} {shade, see in-figure legend)} for the WN signal in the three conditions. The anti-correlation is present in both GAN conditions; the ablation condition remains flat at 0.600 for the entire session.}
\label{fig:c22c33}
\end{figure}

Figures~\ref{fig:delta_c22} and~\ref{fig:c22c33} show, respectively, the comparison of $\Delta c_{22}$ for all signals and the temporal evolution of $c_{22}$ and $c_{33}$ for the WN signal, confirming a consistent positive direction across materials (14/15 \textsc{gan} sessions); per-signal significance is
limited by $n=3$.

\section{Observed Sonic Behavior}\label{sec:sonoro}

The qualitative observations reported are based on the experimental sessions conducted. The sonic behavior depends on the interaction between the input material and the {active source}. {With the \textbf{fixed} source}, the sound remains recognizably tied to the source: the transformation acts on the FDN parameters, modulating timbral coloration, density of resonances, decay times, and energetic distribution, producing audible but non-disintegrating drifts.

{With the \textbf{circular} source}, the rewriting of the buffer leads to a progressive substitution of the original content, analogously to \cite{lucier:1969}, but with a non-deterministic trajectory actively governed by the adversarial mechanism. The parameter $\alpha_{buf}$ directly regulates the ``life'' of these transformations, determining the speed with which the original material evolves toward the product processed by the network.

The nature of the input signal conditions the sonic evolution: materials with defined transient attacks tend to preserve a trace of their rhythmic articulation, until the buffer-rewriting cycle overwrites it completely. By contrast, continuous spectra favor a predominantly timbral and textural evolution, less bound to sharp structural references.
Finally, the \textbf{live} {source}, being buffer-free, offers maximum reactivity, producing an instantaneous adaptive transformation in response to the variations of the external source.

\section{Discussion}\label{sec:discussione}

The empirical results indicate that the adversarial mechanism operates as formally predicted: discriminator and generator interact in a measurable way, maintaining a stable stationary regime ($D \approx 0.44$,  {offset from the} {$D=0.5$ reference value} {associated with the} {classical Nash equilibrium}) {-- an empirically} {stable operating point} {of this specific} {control loop, not} {a formal extension} {of the Nash-equilibrium} {concept itself} and generating differentiated parametric dynamics depending on the input conditions.

The choice of a minimal discriminator (a perceptron with five energetic features) aims at the transparency of the learning process, evaluating the energetic coherence with respect to the initial state (``DNA'') rather than the absolute timbral quality. This modularity makes the architecture extensible: the evaluation function can be replaced by more complex metrics (e.g.\ spectral features or neural representations) without altering the feedback logic of the generator.

The dependence of the ``DNA'' on the initial conditions and the warm-up phase introduces a sensitivity to context which, analogously to the dependence on the acoustics of the physical space in \textit{I Am Sitting in a Room} \cite{lucier:1969}, transforms the deterministic non-repeatability of the individual trajectories into a structural variable of the system, rather than into a limitation.

In this perspective, LETHE exemplifies a distributed agency in algorithmic composition \cite{hayles:1999, lewis:2000}: the composer delimits the space of parametric possibilities, delegating the specific outcome to the emergent interaction of the system. The transparency of the architecture facilitates the concrete exploration of the generative mechanisms, realizing Di Scipio's paradigm \cite{discipio:2003}, in which the audio signal serves simultaneously as output, input, and control interface.

\section{Conclusions}\label{sec:conclusioni}

In this perspective, LETHE does not generate sound material through the symbolic prescription of events, but operates as a dynamical system that explores a parametric space: guided by an internal objective and constrained by a physical structure, it produces non-deterministic evolutions. The system  {adopts the} {vocabulary and structure} {of} an adaptive energy-domain audio GAN which, instead of replicating the original framework, isolates its essential dynamic structure: a real-time self-evaluation mechanism that operates without depending on external datasets.

In this context, agency does not reside in a single actor, but emerges from the \textit{loop} itself, through the continuous interaction among the components constrained by the hierarchical coupling. The architecture lends itself to three application domains: \textbf{compositional}, for the generation of material based on the negotiation between coherence and variation; \textbf{didactic}, thanks to the transparency of an explicit-parameter model; and \textbf{theoretical}, illustrating the applicability of the adversarial approach to minimal real-time systems.

Future directions concern the extension of the evaluation function to spectral metrics or learned representations, and the analysis of stability in long-duration sessions, exploiting the modularity of the architecture. {Further methodological} {controls -- a} {frozen or randomly-initialized} {discriminator, a symmetric,} {unbiased perturbation prior,} {and a random-walk} {baseline under the} {same stability constraints} {-- would help} {isolate the specific} {contribution of the} {adversarial interaction from} {the structural biases} {of the perturbation} {scheme, and are} {left for future} {validation.} {The validation deliberately} {used synthetic test} {signals (white noise,} {impulse train, pink} {noise, square and} {sawtooth waves) rather} {than the human} {voice used by} {Lucier, in order} {to obtain controlled,} {repeatable, spectrally well-defined} {conditions for the} {statistical comparisons of} {Section~6; voice and} {other real-world material} {remain available through} {the \texttt{\textbackslash mic} and} {\texttt{\textbackslash live} acquisition modes} {and are a} {natural direction for} {artistic and perceptual} {follow-up work.} The experimental sessions document the functioning of the mechanism under controlled conditions, while the sonic behavior in application scenarios is discussed in Section~\ref{sec:sonoro}.

\bibliography{lethe_paper_camera_ready}

\end{document}